\documentclass[sigplan]{acmart}
\setcopyright{rightsretained}
\acmConference[TyDe '26]{The ACM SIGPLAN International Workshop on
  Type-Driven Development}{August 26--27, 2026}{Paris, France}
\acmYear{2026}
\usepackage{amsmath}
\usepackage{graphicx}
\usepackage{listings}
\usepackage{xcolor}
\usepackage{url}
\usepackage{tikz}
\usetikzlibrary{automata,positioning,arrows.meta}
\lstdefinelanguage{Agda}{
  morekeywords={data,record,where,field,with,let,in},
  sensitive=true,
  morecomment=[l]{--},
  literate=
    {×}{{$\times$}}1
    {λ}{{$\lambda$}}1
    {ᵥ}{{$_{\mathrm{v}}$}}1
    {₁}{{$_1$}}1
    {₂}{{$_2$}}1
    {₃}{{$_3$}}1
    {→}{{$\to$}}1
    {∀}{{$\forall$}}1
    {∃}{{$\exists$}}1
    {∷}{{$::$}}1
    {≡}{{$\equiv$}}1
    {⊤}{{$\top$}}1
    {⊸}{{$\multimap$}}1
}
\newcommand{\teluguglyph}[2][1.7ex]{
  \raisebox{-0.35ex}{\includegraphics[height=#1]{telugu-#2.pdf}}}
\begin{document}
\title{Type-Driven Tokenization for Brahmic Scripts}
\subtitle{A Pearl}
\author{Sai Hemanth Kapila}
\affiliation{
  \institution{Microsoft}
  \country{India}
}
\email{skapila@microsoft.com}
\orcid{0009-0006-1546-1223}
\author{Rakshika Bagavathy}
\affiliation{
  \institution{Microsoft}
  \country{India}
}
\email{rakb@microsoft.com}
\orcid{0009-0008-4880-0281}
\begin{abstract}
Standard tokenizers used in large language models produce malformed text
when applied to Brahmic scripts. They are a family of abugidas, writing
systems whose consonants carry an inherent vowel that dependent marks
can modify. They include Devanagari, Telugu, Tamil, Kannada, and others.
The underlying issue is that these tokenizers violate
orthographic constraints that do not arise in alphabetic scripts like
English. We observe that while English orthography forms a \emph{semigroup}
(any two valid tokens can be freely concatenated), Brahmic orthography
forms a \emph{partial semigroup}: not every concatenation yields a valid
string. We formalise this distinction in Agda, model valid Brahmic
tokens as chains in a transition system, and derive a provably correct
\texttt{fixToken} function that extends any candidate token to respect
orthographic boundaries.  We then show how this formal derivation translates
into a practical patch for SentencePiece as well as a standalone Rust-based
pre-tokenizer library, eliminating the observed errors across Indic scripts.
\end{abstract}
\maketitle
\section{Introduction}
\label{sec:intro}
Digital text is encoded as a sequence of Unicode \emph{code points},
but a code point need not correspond to a complete written character.
A single grapheme, or orthographic unit, may comprise a base character
and one or more combining marks, while a word may comprise several such
units. Text-processing systems rarely operate directly on these raw
code points. Search engines, spell checkers, machine-translation
systems, and, most prominently today, large language models (LLMs)
rely on a process called \emph{tokenization}. which segments text
into text into reusable units called \emph{tokens}. In this paper,
tokens are \emph{subwords}, vocabulary entries that may contain several
code points but may be shorter than a word. A token is therefore atomic
to the model, even though it may have internal Unicode and orthographic
structure.
Modern systems learn their inventory of tokens, their \emph{vocabulary},
from corpus statistics. The two dominant
algorithms, Byte Pair Encoding (BPE)~\cite{BPE} and
Unigram~\cite{Unigram}, used in GPT- and Llama-style architectures,
build vocabularies of frequently co-occurring code-point sequences
and then split any input into a sequence of vocabulary entries. Such
\emph{frequency-driven subword tokenizers}
know nothing about the writing system they operate on. Any
sufficiently frequent sequence of code points can become a token,
and a token boundary may fall between any two code points.
For English this indifference is harmless, because every letter is a
self-contained symbol. In other words, any fragment of valid text is itself
displayable text. However, the same cannot be said about the Brahmic scripts of
South and South-East Asia. These scripts are
\emph{abugidas}~\cite{Bright99}, where each
consonant letter carries an inherent vowel (Telugu \teluguglyph{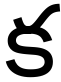}
reads \emph{ka}), and other vowels are written as \emph{dependent}
marks attached to a consonant rather than as free-standing letters.
A script's \emph{orthography} determines which character sequences
constitute valid written text and therefore constrains which code
points may occur in succession. Standard
``out-of-the-box'' tokenizers routinely violate these constraints.
Our experiments training LLMs on Telugu, Tamil, Devanagari, and
Gujarati corpora show frequent overlapping characters, unreadable
sequences, and isolated dependent modifiers rendered as empty dotted
circles (e.g., \teluguglyph{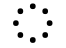}).
Recent empirical studies confirm that this category of errors is
widespread across complex writing systems. Manodnya and
Giri~\cite{OSPE} showed that treating \emph{orthographic
syllables} as the atomic unit for Indic
language modelling yields a 30\% improvement in compression ratio
over BPE and significantly better language-model performance. This is primarily
because BPE produces tokens that are not valid character sequences
in abugida scripts. De Nardi and
Manodnya~\cite{OrthographicStructureMatters} extended this finding
to Arabic-script languages in a multilingual evaluation.
The problem is therefore well-established. Frequency-driven subword tokenizers
violate orthographic constraints  that do not arise in languages like English.
\emph{What has been missing is a formal account of why these violations occur and
a provably correct fix.}
In this pearl, we supply that account. We observe that the distinction
is \emph{algebraic}. English orthography, viewed as an operation on
character sequences, forms a \emph{semigroup}, where concatenation of any
two valid tokens always yields a valid token. Brahmic scripts
(Devanagari, Telugu, Tamil, Kannada, and others) contain \emph{dependent characters}
like vowel signs and vowel suffixes, that are meaningful only when attached to
a preceding base character. Other orthographies, such as Thai or
Vietnamese in decomposed Unicode form, show similar properties. These orthographies form a \emph{partial
semigroup}, where concatenation may fail when the boundary between two
sequences violates a transition constraint. Tokenizers designed for
semigroups produce invalid output when applied to partial semigroups.
Although we quantify the damage with language-modelling metrics, the
constraint itself is not an artifact of LLMs. Rather it is a fact about the
computational representation of writing systems. It applies to any
system that splits Unicode text, including a renderer, search index,
or tokenizer.
We formalise this distinction in Agda and derive a provably correct
token-boundary fix:
\begin{enumerate}
  \item We model orthographic structure at two algebraic levels: total
        and partial semigroups, in Agda (Section~\ref{sec:orthography}).
  \item We encode Brahmic orthography as a typed transition system. We then
        define valid tokens as chains in this system, and derive a
        \texttt{fixToken} function with machine-checked proofs of
        validity preservation, remainder correctness, and completeness
        (Section~\ref{sec:formal}).
  \item We prove that Brahmic script is a partial semigroup, and to illustrate
        that the pattern applies beyond this family, we provide a proof for Vietnamese too.
        (Section~\ref{sec:formal}).
  \item We translate the formal derivation into a patch for
      SentencePiece, Kudo and Richardson's open-source library for
      training and applying BPE and Unigram tokenizers~\cite{SentencePiece}.
      We also provide a configurable Rust pre-tokenizer library, eliminating observed errors in
        Telugu, Tamil, Hindi, and Gujarati
        (Section~\ref{sec:implementation}). At the cost of a modest
        (about 10\%) increase in token count, the resulting
        tokenization cuts language-model perplexity nearly in half
        (from 166 to 85 for Unigram, and from 159 to 92 for BPE) on a
        Telugu corpus (Section~\ref{sec:evaluation}).
\end{enumerate}
\section{Orthography as Algebraic Structure}
\label{sec:orthography}
This section first reviews the character classes that make up Brahmic
scripts, and shows with a concrete Telugu example, how a badly placed
token boundary produces malformed text. It then captures the
difference between alphabetic and Brahmic orthographies as two
algebraic structures in Agda, a semigroup and a partial semigroup.
\subsection{Brahmic Script Characteristics}
Brahmic scripts are abugidas where each consonant letter inherently carries a
vowel (typically `a'). The character classes relevant to tokenization
are:
\begin{itemize}
  \item \textbf{Consonant symbols}: carry an inherent vowel
        (e.g., Telugu \textit{ka} = U+0C15).
  \item \textbf{Independent vowels}: stand-alone vowel letters
        (e.g., Telugu \textit{a} = U+0C05).
  \item \textbf{Dependent vowels} (matras): modify a consonant's inherent
        vowel. These cannot appear independently
        (e.g., Telugu \textit{-i} = U+0C3F).
  \item \textbf{Virama} (halant): cancels the inherent vowel, creating a
        \emph{dead} consonant that joins with the next
        (e.g., U+0C4D).
  \item \textbf{Vowel suffixes}: nasalisation or aspiration markers that follow a vowel
        (e.g., Telugu anusvara = U+0C02).
\end{itemize}
\paragraph{A concrete example.}
Consider the Telugu word \emph{vij\~{n}a} (a prefix meaning
``knowledge''), encoded as the Unicode sequence in
Table~\ref{tab:telugu-example}.
\begin{table}[h]
\centering
\small
\begin{tabular}{clll}
\hline
\textbf{Codepoint} & \textbf{Glyph} & \textbf{Role} & \textbf{Can start?} \\
\hline
U+0C35 & \teluguglyph{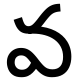}      & Consonant       & Yes \\
U+0C3F & \teluguglyph{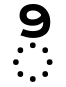}  & Dependent vowel & \textbf{No} \\
U+0C1C & \teluguglyph{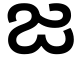}      & Consonant       & Yes \\
U+0C4D & \teluguglyph{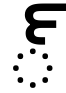}  & Virama          & \textbf{No} \\
U+0C1E & \teluguglyph{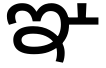}     & Consonant       & Yes \\
U+0C3E & \teluguglyph{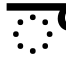} & Dependent vowel & \textbf{No} \\
\hline
\end{tabular}
\caption{Decomposition of the Telugu syllable cluster
\teluguglyph[2.1ex]{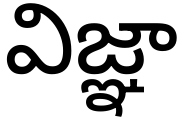} (\emph{vij\~{n}\={a}}).
Characters marked \textbf{No} are dependent, they cannot begin a
valid token.}
\label{tab:telugu-example}
\end{table}
If a tokenizer splits this sequence after position~3
(U+0C1C, the consonant \emph{ja}), the right fragment begins with the
virama U+0C4D. If an LLM later emits such a token after anything other
than a consonant, the result is ill-formed.  A rendering engine displays
this as a dotted circle (\teluguglyph{dotted-circle}) followed by the virama glyph,
producing visually corrupted output. Valid tokens must begin with a consonant or independent vowel.
\subsection{Two Levels of Orthographic Structure}
We model orthography at two algebraic levels in Agda.
\paragraph{Level 1: Semigroup (English, German).}
Any two valid tokens can always be combined and the combination operation is associative.
\begin{lstlisting}[language=Agda]
record Orthography₁ (Token : Set)
    : Set where
  field
    combine : Token → Token → Token
    assoc   : ∀ (x y z : Token)
      → combine (combine x y) z
      ≡ combine x (combine y z)
\end{lstlisting}
\paragraph{Level 2: Partial Semigroup.}
In scripts such as Brahmic and Vietnamese, combination may fail when
tokens are orthographically incompatible.
\begin{lstlisting}[language=Agda]
record Orthography₂ (Token : Set)
    : Set where
  field
    combine : Token → Token → Maybe Token
    assoc   : ∀ (x y z : Token)
              (xy xyz : Token)
      → combine x y ≡ just xy
      → combine xy z ≡ just xyz
      → ∃ λ yz
        → combine y z ≡ just yz
        × combine x yz ≡ just xyz
\end{lstlisting}
BPE and Unigram tokenizers implicitly assume the total structure
Orthography$_1$. For scripts living in Orthography$_2$, we need a
tokenizer that respects the partial structure.
\section{Formal Model of Brahmic Tokenization}
\label{sec:formal}
We now make the picture of Section~\ref{sec:orthography} precise. We
encode the Brahmic character classes and their permitted transitions
as Agda datatypes, define valid tokens as chains in the resulting
transition system, and derive a boundary-repairing function
\texttt{fixToken} together with machine-checked proofs of its
correctness. We then replay the same construction for Vietnamese and
close by exhibiting both scripts as instances of the partial-semigroup
record.
\subsection{Character Types and Transitions}
\begin{figure}[h]
\centering
\begin{tikzpicture}[
  ->,>=Stealth,
  node distance=2.2cm,
  state/.style={rectangle, draw, rounded corners, minimum width=1.6cm,
                minimum height=0.7cm, font=\small},
  every edge/.style={draw, font=\scriptsize}
]
  \node[state] (C)  {Consonant};
  \node[state, right=of C] (V)  {Indep.\ Vowel};
  \node[state, below=of C] (DV) {Dep.\ Vowel};
  \node[state, below=of V] (VS) {Vowel Suffix};
  \node[state, below right=1.4cm and 0.5cm of C] (Vi) {Virama};
  \path (C)  edge [bend left=15] node[above] {} (V);
  \path (C)  edge node[left] {} (DV);
  \path (C)  edge node[below left] {} (Vi);
  \path (C)  edge node[above right] {} (VS);
  \path (C)  edge [loop above] node {} (C);
  \path (V)  edge [bend left=15] node[below] {} (C);
  \path (V)  edge [loop above] node {} (V);
  \path (V)  edge node[right] {} (VS);
  \path (DV) edge [bend right=15] node[below] {} (C);
  \path (DV) edge [bend left=10] node[below] {} (V);
  \path (DV) edge node[below] {} (VS);
  \path (Vi) edge [bend left=10] node[right] {} (C);
  \path (Vi) edge [bend right=25] node[below right] {} (V);
  \path (VS) edge [bend right=20] node[right] {} (C);
  \path (VS) edge [bend left=15] node[above] {} (V);
\end{tikzpicture}
\caption{Valid transitions between Brahmic character classes.
Edges represent the constructor names in the Agda formalisation
(e.g., the edge Consonant $\to$ Virama corresponds to
\texttt{DeadConsonant}).}
\label{fig:transitions}
\end{figure}
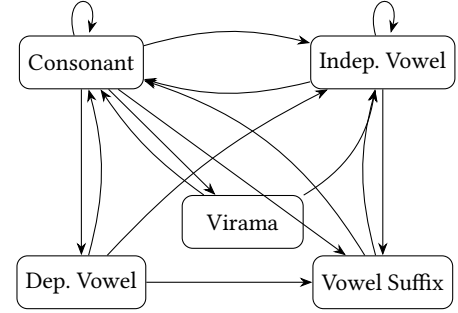

We model Brahmic character categories as a finite type and valid
transitions as a relation. Figure~\ref{fig:transitions} shows the
permitted transitions between character classes. An edge from $A$ to
$B$ means a character of class $B$ may immediately follow one of class
$A$ within a valid token.
\par\medskip
\noindent\begin{minipage}{\columnwidth}
\begin{lstlisting}[language=Agda]
data UnicodeBrahmic : Set where
  Consonant        : UnicodeBrahmic
  IndependentVowel : UnicodeBrahmic
  DependentVowel   : UnicodeBrahmic
  VowelSuffix      : UnicodeBrahmic
  Virama           : UnicodeBrahmic
data _⊸_ : UnicodeBrahmic
          → UnicodeBrahmic → Set where
  DeadConsonant : Consonant ⊸ Virama
  VowelSymbol   : Consonant ⊸ DependentVowel
  NextVowel     : ∀ {t} → t ⊸ IndependentVowel
  NextConsonant : ∀ {t} → t ⊸ Consonant
  VowelSuffix₁  :
    Consonant ⊸ VowelSuffix
  VowelSuffix₂  :
    IndependentVowel ⊸ VowelSuffix
  VowelSuffix₃  :
    DependentVowel ⊸ VowelSuffix
\end{lstlisting}
\end{minipage}
The transition relation \(\mathord{\_\!\multimap\!\_}\) encodes the state machine of
valid character sequences. A valid token is a chain of transitions
starting from a consonant or independent vowel.
\subsection{Valid Chains and Tokens}
A \texttt{ValidChain} witnesses that every adjacent pair in a character
sequence satisfies the transition relation. It is indexed by the
\emph{last} element of the chain, which makes appending two chains
type-safe, since we can verify that the last element of the first chain
can transition to the first element of the second.
\begin{lstlisting}[language=Agda]
data ValidChain : UnicodeBrahmic
    → List UnicodeBrahmic → Set where
  base : ∀ {t} → ValidChain t (t ∷ [])
  step : ∀ {t₁ t₂ last}
         {rest : List UnicodeBrahmic}
       → (t₁ ⊸ t₂)
       → ValidChain last (t₂ ∷ rest)
       → ValidChain last (t₁ ∷ t₂ ∷ rest)
\end{lstlisting}
\noindent The \texttt{base} constructor witnesses a single-character
chain. The \texttt{step} prepends a character by supplying a transition
proof.\\
A \texttt{ValidBrahmicToken} is then a chain that starts at a
permitted initial state.
\par\medskip
\noindent\begin{minipage}{\columnwidth}
\begin{lstlisting}[language=Agda]
data ValidBrahmicToken
    : List UnicodeBrahmic → Set where
  consonant-start : ∀ {last} {rest}
    → ValidChain last (Consonant ∷ rest)
    → ValidBrahmicToken (Consonant ∷ rest)
  vowel-start : ∀ {last} {rest}
    → ValidChain last
        (IndependentVowel ∷ rest)
    → ValidBrahmicToken
        (IndependentVowel ∷ rest)
\end{lstlisting}
\end{minipage}
\noindent Crucially, there is no constructor for chains beginning with
a dependent vowel, virama, or vowel suffix. These are ruled out
\emph{by construction}. A dependent character can never begin a valid
token.
A key closure property follows: any two valid tokens can be
concatenated to form a valid token (via \texttt{NextConsonant} or
\texttt{NextVowel} at the boundary). This asserts that the set of
valid tokens forms a semigroup under concatenation. This closure property does not contradict our partial-semigroup characterization. Concatenation is total on the restricted set of already-valid tokens, because the second token necessarily begins with a consonant or independent vowel. Partiality arises on the larger domain of candidate fragments considered by a tokenizer, a fragment may begin with a dependent character, and concatenating it is defined only when the boundary transition is permitted. Note that valid tokens are inherently non-empty (both constructors require at least one character). There is no unit element, which is why the appropriate
algebraic structure is a semigroup rather than a monoid.
\subsection{The \texttt{fixToken} Function}
To see where types must meet practice, recall that a subword tokenizer
processes text from left to right, selecting a vocabulary entry that
matches a prefix of the remaining input, emitting it, and continuing
with the rest of the text. BPE determines this selection through greedy
merging, whereas Unigram uses a likelihood criterion. The right boundary
of every emitted token is thus chosen by frequency, with no regard for
the script. The specification developed above can be stated in one
sentence: \emph{never place a token boundary where the fragment to
its right would begin with a dependent character.}
At the heart of our approach is a recursive function,
\texttt{fixToken}, that repairs a boundary proposed by a standard
tokenizer so that it meets this specification. It is designed to
wrap an existing BPE or Unigram tokenizer as a post-processing step.
Given a candidate token (already identified by
BPE or Unigram) and the remaining input, the function performs a one-character
lookahead. If the next character is a valid token start (a consonant
or independent vowel), the candidate is complete and returned as is.
Otherwise, the next character is a dependent continuation that
\emph{must} be absorbed into the current token, so append it and recur.
\begin{lstlisting}[language=Agda]
fixToken : List UnicodeBrahmic
  → List UnicodeBrahmic
  → List UnicodeBrahmic
  × List UnicodeBrahmic
fixToken cand [] = cand , []
fixToken cand (Consonant ∷ r)
  = cand , Consonant ∷ r
fixToken cand (IndependentVowel ∷ r)
  = cand , IndependentVowel ∷ r
fixToken cand (DependentVowel ∷ r)
  = fixToken
      (cand ++ DependentVowel ∷ []) r
fixToken cand (VowelSuffix ∷ r)
  = fixToken (cand ++ VowelSuffix ∷ []) r
fixToken cand (Virama ∷ r)
  = fixToken (cand ++ Virama ∷ []) r
\end{lstlisting}
\paragraph{Repairing the Telugu example.}
Consider again the split from Table~\ref{tab:telugu-example}. The tokenizer
has proposed the candidate U+0C35 U+0C3F U+0C1C, whose character
classes are Consonant, DependentVowel, and Consonant. The
remainder is U+0C4D U+0C1E U+0C3E, classified as Virama, Consonant,
and DependentVowel. On its first step, \texttt{fixToken} observes that
the remainder begins with a virama, which cannot start a valid token.
It therefore appends U+0C4D to the candidate and recurses on the
remaining U+0C1E U+0C3E. The second step sees that U+0C1E is a
consonant and hence a valid token start, so the recursion stops. The
function returns U+0C35 U+0C3F U+0C1C U+0C4D as the repaired token and
U+0C1E U+0C3E as the remainder. Thus the boundary moves one code point
to the right, no character is lost, and the remainder now begins at a
valid boundary.
\noindent The function terminates because the remainder shrinks by one
element at each recursive call. It returns a pair: the extended token
and the unconsumed remainder. The lookahead is the pattern match on the
head of the second argument. There is no need for  backtracking or unbounded search.
\subsection{Correctness Properties}
We prove three key properties that together guarantee that
\texttt{fixToken} is a safe, lossless post-processing step:
\paragraph{Validity preservation.} The lemma establishes that
 token validity is preserved on well-formed input. If the candidate is a valid token
and the full concatenation (candidate appended with remainder) is also
valid, then the extended token returned by \texttt{fixToken} is valid:
\par\medskip
\noindent\begin{minipage}{\columnwidth}
\begin{lstlisting}[language=Agda]
fixTokenTokenValid : ∀ {l₁ l₂}
  → ValidBrahmicToken l₁
  → ValidBrahmicToken (l₁ ++ l₂)
  → let (parsed , _) = fixToken l₁ l₂
    in  ValidBrahmicToken parsed
\end{lstlisting}
\end{minipage}
\paragraph{Remainder validity.} The remainder always begins with a
valid start character (or is empty), ensuring that the \emph{next}
token can also be validly formed. The predicate \texttt{RemainderOk}
states exactly this, reusing the \texttt{IsValidBrahmicStart} type
that singles out the two permitted initial states:
\begin{lstlisting}[language=Agda]
data IsValidBrahmicStart
    : UnicodeBrahmic → Set where
  ConsonantStart :
    IsValidBrahmicStart Consonant
  IndependentVowelStart :
    IsValidBrahmicStart IndependentVowel
RemainderOk
    : List UnicodeBrahmic → Set
RemainderOk [] = ⊤
RemainderOk (x ∷ _) =
  IsValidBrahmicStart x
fixTokenRemainderOk : ∀ (l₁ l₂)
  → let (_ , rem) = fixToken l₁ l₂
    in  RemainderOk rem
\end{lstlisting}
\paragraph{Completeness.} This lemma asserts that no characters are lost or duplicated. The
extended token concatenated with the remainder equals the original
input:
\begin{lstlisting}[language=Agda]
fixTokenComplete : ∀ (l₁ l₂)
  → let (parsed , rem) = fixToken l₁ l₂
    in  parsed ++ rem ≡ l₁ ++ l₂
\end{lstlisting}
\noindent Together, these three properties ensure that
\texttt{fixToken}, applied as a post-processing step after any
standard tokenizer, produces a valid tokenization. Every token respects
orthographic constraints and the original
text is faithfully preserved.
\subsection{Vietnamese: A Simpler Instance}
The same pattern applies to other partial-semigroup orthographies.
Vietnamese in decomposed Unicode form provides a second
example. The orthography is simpler than Brahmic (three character classes instead of
five). It is also partial.
The orthographic constraint in Vietnamese is that a base character
may carry at most one diacritic (circumflex, breve, horn) followed by
at most one tone mark (acute, grave, hook, tilde, dot below). Stacking
two diacritics, two tones, or placing a diacritic after a tone are all
invalid.
\par\medskip
\noindent\begin{minipage}{\columnwidth}
\begin{lstlisting}[language=Agda]
data Vietnamese : Set where
  BaseCharacterᵥ : Vietnamese
  Diacriticᵥ     : Vietnamese
  Toneᵥ          : Vietnamese
data _⊸ᵥ_ : Vietnamese
           → Vietnamese → Set where
  Markᵥ :
    BaseCharacterᵥ ⊸ᵥ Diacriticᵥ
  ToneAfterBaseᵥ :
    BaseCharacterᵥ ⊸ᵥ Toneᵥ
  ToneAfterDiacriticᵥ :
    Diacriticᵥ ⊸ᵥ Toneᵥ
  NextBaseᵥ :
    ∀ {t} → t ⊸ᵥ BaseCharacterᵥ
\end{lstlisting}
\end{minipage}
\noindent The invalid transitions (Diacritic $\to$ Diacritic,
Tone $\to$ Tone, Tone $\to$ Diacritic) have no
constructors, and are thus ruled out by the type, as before.
The \texttt{fixToken} function follows the identical structure:
\begin{lstlisting}[language=Agda]
fixTokenᵥ cand [] = cand , []
fixTokenᵥ cand (BaseCharacterᵥ ∷ r)
  = cand , (BaseCharacterᵥ ∷ r)
fixTokenᵥ cand (Diacriticᵥ ∷ r)
  = fixTokenᵥ (cand ++ Diacriticᵥ ∷ []) r
fixTokenᵥ cand (Toneᵥ ∷ r)
  = fixTokenᵥ (cand ++ Toneᵥ ∷ []) r
\end{lstlisting}
\noindent The same correctness properties (validity preservation,
remainder correctness, completeness) hold and are proved analogously.
\subsection{Instances of Orthography\texorpdfstring{$_2$}{2}}
We close the formal development by constructing concrete
\texttt{Orthography$_2$} instances for both scripts. For each, we
define a decidable boundary-checking function \texttt{canFollow} that
returns \texttt{true} when the transition relation is inhabited,
and a \texttt{combine} that concatenates two character lists only when
the boundary is valid:
\begin{lstlisting}[language=Agda]
combineBrahmic : List UnicodeBrahmic
  → List UnicodeBrahmic
  → Maybe (List UnicodeBrahmic)
combineBrahmic (x ∷ xs) (y ∷ ys)
  with canFollow (getLast x xs) y
... | true  = just (x ∷ xs ++ y ∷ ys)
... | false = nothing
\end{lstlisting}
\noindent We prove that \texttt{canFollow} is both sound and
complete with respect to \(\mathord{\_\!\multimap\!\_}\), and that
\texttt{combineBrahmic} satisfies associativity, yielding:
\begin{lstlisting}[language=Agda]
brahmicOrthography₂ :
  Orthography₂ (List UnicodeBrahmic)
vietnameseOrthography₂ :
  Orthography₂ (List Vietnamese)
\end{lstlisting}
\noindent This completes the bridge between the abstract algebraic
structure (Section~\ref{sec:orthography}) and the concrete
transition-based model. Brahmic and Vietnamese character sequences
 are certified instances of the partial semigroup record.
\paragraph{Artifact availability.}
The artifact accompanying this paper with complete Agda formalisation used in this work, is available at
\url{https://github.com/rakshikab/brahmi_script_formalization/blob/main/formalization/Brahmic.agda}.
The development type-checks under the \texttt{--safe} flag with
Agda~2.9.0 and version~2.3 of the Agda standard library.
\section{Implementation}
\label{sec:implementation}
The formal model of Section~\ref{sec:formal} gives us a clear
specification. We never allow a token boundary where the right fragment
would begin with a dependent character. We now describe two practical
realisations of this specification.
\subsection{The Dual Problem of Pre-tokenizers}
Tokenizer libraries distinguish the tokenizer proper, which learns
and applies the subword vocabulary, from a \emph{pre-tokenizer}: a
preliminary pass that carves the input into coarse fragments,
classically at whitespace and punctuation. The vocabulary is then
learned and applied strictly \emph{within} fragments, with no token ever
crossing a fragment boundary.
SentencePiece~\cite{SentencePiece}, the widely used open-source
tokenizer library that implements both BPE and Unigram, exposes this
mechanism for extension via its
\texttt{Pretokenizer\allowbreak For\allowbreak Training\allowbreak Interface}. Users can supply custom
logic that forces the tokenizer to \emph{break} at specified positions.
This solves the
\emph{dual} of our problem. Pre-tokenizers specify where tokens
\emph{must} be split. However, we need to specify where tokens \emph{must not}
be split. No existing SentencePiece interface supports the latter
constraint directly.
\subsection{SentencePiece Patch}
The most direct translation of our formal result is a filter in
SentencePiece's \texttt{IsValidSentencePiece()} function \\
(\texttt{src/trainer\_interface.cc}), which validates every candidate
token during training. We add the following predicate that rejects any candidate
whose first character is a dependent or combining mark.
\begin{lstlisting}[language=C++]
static bool IsDependentOrCombining(char32 c) {
  // Telugu dependent vowels + virama
  if (c >= 0x0C3E && c <= 0x0C56) return true;
  if (c == 0x0C4D) return true;
  // Combining diacriticals (Vietnamese)
  if (c >= 0x0300 && c <= 0x036F) return true;
  return false;
}
\end{lstlisting}
\noindent This is the \texttt{canFollow} check from our Agda
development, specialised to the ``is this a valid token start?''
question. By rejecting such candidates at training time, the learned
vocabulary contains only tokens that begin at valid boundaries, implying that the
\texttt{fixToken} lookahead is then never needed at inference time.
Note, however, that this patch hard-codes specific Unicode ranges for
Telugu and Vietnamese. Supporting additional Brahmic scripts (Tamil,
Kannada, Devanagari, etc.) would require extending the range table
for each script. Keeping this maintenance burden in mind, we discuss
a more general solution below.
\subsection{Rust Pre-tokenizer Library}
While the SentencePiece patch works for specific hard-coded Unicode
ranges, we wanted a more flexible solution configurable for arbitrary
Brahmic scripts without modifying SentencePiece's source. We built a
standalone Rust library that:
\begin{enumerate}
  \item Pre-tokenizes input text by identifying orthographically valid
        grapheme clusters using the transition rules (the Rust
        analogue of \texttt{fixToken}).
  \item Maps each cluster to an unused Unicode code point from the
        Private Use Area (PUA), producing a one-to-one encoding where
        each PUA symbol represents an indivisible orthographic unit.
  \item Passes the transformed text to SentencePiece, which now
        operates on atomic symbols that cannot be split incorrectly.
  \item Reverses the mapping when consuming LLM output, recovering
        valid Brahmic text.
\end{enumerate}
This approach has been tested with Telugu, Tamil, Gujarati, and
Devanagari, eliminating orthographic errors  across all
tested configurations.
The boundary logic at the core of the library is a direct, if
manual, transcription of the Agda development: a Rust \texttt{enum}
mirrors \texttt{UnicodeBrahmic}, and a case analysis on the incoming
character's class, guarded by the class of the preceding symbol,
makes exactly the accept/reject decisions of \texttt{canFollow},
clause for clause. The code surrounding that core (tables mapping
Unicode ranges to character classes for each script, PUA bookkeeping,
and the SentencePiece plumbing) is routine engineering with no formal
counterpart.
The key insight connecting the implementation back to our formal
development is that standard tokenizers cannot subdivide a single
Unicode code point. This forces the language model to treat our
\texttt{ValidBrahmicToken}s as indivisible units. The PUA encoding
is the runtime mechanism that enforces what the type system guarantees
statically that every token respects orthographic boundaries.
\section{Evaluation}
\label{sec:evaluation}
A natural concern is whether enforcing orthographic correctness
degrades tokenizer efficiency. We evaluate the Rust pre-tokenizer of
Section~\ref{sec:implementation} on two metrics, one measuring cost
and one measuring benefit. \emph{Normalized Sentence Length}
(NSL)~\cite{dagan2024} is the ratio of the token count our tokenizer
produces for a sentence to the token count of a reference tokenizer
(here, GPT-4o's). It captures efficiency, since a model must process
and generate more tokens for the same text when NSL rises.
\emph{Perplexity} measures downstream language-model quality, how well
a model trained on the resulting tokens predicts held-out text, with
lower values indicating better predictions.
\paragraph{Setup.}
We trained SentencePiece Unigram and BPE tokenizers, each with a
vocabulary size of 8000, on 75K Telugu Wikipedia articles (12.1M
tokens). For each tokenization configuration, we then trained
nanoGPT~\cite{NanoGPT}, a minimal implementation of GPT-2, on the same
corpus and measured perplexity on a held-out set of 1M tokens of Telugu
news text~\cite{telugu_news_dataset}.
\paragraph{NSL}
The Brahmi pre-tokenizer increases NSL marginally, from 0.65
to 0.71 (Unigram) and 0.64 to 0.72 (BPE), remaining well below
the GPT-4o baseline of 1.0. The small increase reflects the fact
that our approach sometimes prevents merges that cross orthographic
boundaries, producing slightly more tokens per sentence. This cost
is inherent in the specification, not in the transcription from Agda
to Rust: NSL counts tokens rather than running time, and any
tokenizer that refuses to split orthographic units must occasionally
spend more tokens on the same text, however it is implemented.
\paragraph{Perplexity.}
The effect on language-model quality is dramatic and positive.
Perplexity drops from 166 to \textbf{85} (Unigram) and from 159 to
\textbf{92} (BPE), a reduction of nearly 50\%. Because each token
now corresponds to a linguistically meaningful unit, the model can
learn more coherent representations and predict subsequent tokens
far more accurately.
\paragraph{Summary.}
A marginal 10\% increase in token count buys a 50\% reduction in
perplexity. Orthographic correctness improves downstream model performance.
The formal guarantees from Section~\ref{sec:formal} translate
directly into measurable empirical gains.
\section{Related Work}
\label{sec:related}
\paragraph{Tokenization for non-Latin scripts.}
{\sloppy
Manodnya and Giri~\cite{OSPE} introduced Orthographic Syllable Pair
Encoding (OSPE), which uses orthographic syllables as the atomic
subword unit for Indic languages, achieving 30\% better compression
than BPE and improved perplexity. De~Nardi and
Manodnya~\cite{OrthographicStructureMatters} extended this to
Arabic-script languages in a multilingual evaluation, showing that
BPE-induced fragmentation causes up to 27-point F1 drops and
unstable training. Both works demonstrate the empirical severity of
the problem; our contribution is to explain \emph{why} it arises
(the semigroup vs.\ partial semigroup distinction) and to provide a
machine-checked formal fix.\par}
\paragraph{Script-aware segmentation before LLMs.}
Segmenting Indic text at orthographic rather than statistical
boundaries predates neural language models. Kunchukuttan and
Bhattacharyya~\cite{OrthographicSyllableSMT} used orthographic
syllables as the unit of translation for statistical machine
translation between related languages, outperforming word-,
morpheme-, and character-level units. Mainstream tokenizer libraries
also ship script-aware heuristics: SentencePiece can forbid merges
that span distinct Unicode scripts, and subword training is usually
preceded by regex-based pre-tokenizers keyed on Unicode character
categories. These mechanisms prevent some malformed tokens, but they
do not model the \emph{within-script} dependency constraints that
Brahmic orthography imposes, and they offer no correctness
guarantees. Our contribution is accordingly not new boundary
rules---script-aware segmenters embody the same linguistic
facts---but their formulation as types, which turns ``the tokenizer
respects the script'' from an empirical observation into a
machine-checked property.
\paragraph{Types for parsing and for language.}
Dependently typed and certified parsing is a well-explored area:
Brink et al.~\cite{DependentlyTypedGrammars} embed grammars in Agda
with types that keep semantic actions consistent with the grammar;
Danielsson~\cite{TotalParserCombinators} gives parser combinators
that are total by construction; Sarracino et
al.~\cite{CertifiedDependentRegular} verify, in Coq, parsers for
regular grammars extended with data dependency. These works certify
parsers for programming languages and machine-oriented formats. We
apply the same discipline one level lower, to the writing system
itself. A separate tradition applies type-theoretic tools to natural
language: Lambek's syntactic calculus types sentence
structure~\cite{Lambek}, Barker and Shan analyse quantification and
scope through continuations~\cite{BarkerShan}, and Kovalev and
Angiuli recently formalised a dependently typed calculus of event
structure in Agda~\cite{TelicityCulminativity}. Those works assign
types to syntax and semantics. To our knowledge, ours is the first
dependently typed account of \emph{orthographic} well-formedness.
\paragraph{Subword tokenization.}
BPE~\cite{BPE} and Unigram~\cite{Unigram} are the dominant subword
algorithms, implemented in SentencePiece~\cite{SentencePiece}. Dagan
et al.~\cite{dagan2024} study how to get the most out of tokenizers for
pre-training and domain adaptation but do not address orthographic
validity. None of these works provide formal guarantees about the
well-formedness of learned tokens.
\section{Conclusion}
\label{sec:conclusion}
Recent work~\cite{OSPE,OrthographicStructureMatters} has established
that frequency-driven tokenizers systematically fail on Brahmic and
Arabic-script languages. We have shown that this failure has an
algebraic explanation: these orthographies form partial semigroups,
not semigroups, and tokenizers designed for the latter inevitably violate the
constraints of the former.
By formalising this distinction in Agda, we derived a provably
correct token-fixing function, with machine-checked guarantees of
validity preservation, remainder correctness, and completeness. This
directly translates into practical tokenizer improvements. The same
algebraic pattern applies to Vietnamese and potentially any script
with combining or dependent marks. We hope this pearl illustrates
that type-driven formalisation can both \emph{explain} an empirically
observed failure and deliver a repair whose correctness is
machine-checked rather than merely tested.
\bibliographystyle{ACM-Reference-Format}
\bibliography{references}
\end{document}